%% file: paper.tex
\relax
\documentclass[letterpaper]{article} 
\usepackage{jmlr2e}
\usepackage{times}  
\usepackage{graphicx} 
\usepackage{helvet} 
\usepackage{courier}  
\usepackage{hyperref}

\usepackage{caption} 

\usepackage[table]{xcolor}
\newcommand{\name}{EpiCast~}
\usepackage{arydshln}
\usepackage{graphicx,amsmath,amssymb} 

\usepackage{algorithm}
\usepackage[noend]{algpseudocode}
\usepackage[switch]{lineno}
\title{Accurate Calibration of Agent-based Epidemiological Models\\ with Neural Network Surrogates}
\author{
	\name Rushil Anirudh \email anirudh1@llnl.gov \\
	\addr Lawrence Livermore National Laboratory\\
	Livermore, CA,USA
	\AND
	\name Jayaraman J. Thiagarajan \email jjayaram@llnl.gov \\
	\addr Lawrence Livermore National Laboratory\\
	Livermore, CA,USA
	\AND
	\name Peer-Timo Bremer \email bremer5@llnl.gov \\
	\addr Lawrence Livermore National Laboratory\\
	Livermore, CA,USA
	\AND
	\name  Timothy C. Germann \email tcg@lanl.gov \\
	\addr Los Alamos National Laboratory\\
	Los Alamos, NM, USA
	\AND
	\name  Sara Y. Del Valle \email sdelvall@lanl.gov \\
	\addr Los Alamos National Laboratory\\
	Los Alamos, NM, USA
	\AND
	\name  Frederick H. Streitz \email streitz1@llnl.gov \\
	\addr Lawrence Livermore National Laboratory\\
	Livermore, CA,USA}
\begin{document}

\maketitle

\begin{abstract}
Calibrating complex epidemiological models to observed data is a crucial step to provide both insights into the current disease dynamics, i.e.\ by estimating a reproductive number, as well as to provide reliable forecasts and scenario explorations.
Here we present a new approach to calibrate an agent-based model -- \name{} -- using a large set of simulation ensembles for different major metropolitan areas of the United States.
In particular, we propose: a new neural network based surrogate model able to simultaneously emulate all different locations; and a novel posterior estimation that provides not only more accurate posterior estimates of all parameters but enables the joint fitting of global parameters across regions.

\end{abstract}
\section{Introduction}
\input{intro.tex}

\section{Background}
\input{background.tex}
\section{Proposed Methodology}
\input{proposed.tex}
\section{Experiments and Results}
\input{expts.tex}
\section{Related Work}
\input{related.tex}
\section{Discussion}
\input{discussion.tex}

\section{Acknowledgements}
This work was performed under the auspices of the U.S. Department of Energy by Lawrence Livermore National Laboratory under Contract DE-AC52-07NA27344. Research was supported by the DOE Office of Science through the National Virtual Biotechnology Laboratory, a consortium of DOE national laboratories focused on response to COVID-19, with funding provided by the Coronavirus CARES Act.

\bibliographystyle{ieee}
\bibliography{refs}
\end{document}

%% file: intro.tex
Epidemiological models are playing a key role in the national response to COVID-19:
Model parameters can provide intuitive metrics on disease progression, i.e. by computing an effective reproductive number, model forecasts represent a crucial planning tool, and exploring what-if scenarios can provide insights into the relative trade-off, for example, between social distancing measures and economic impacts.
However, as with all computational models, all such benefits depend on the model parameters being well calibrated to the observed data and the model being generalizable to unknown situations.
Generically, this describes a standard inverse problem of fitting model parameters, under the key assumption that the epidemiological model can effectively describe the observed data.
In practice, epidemiological models face a number of additional challenges that significantly complicate the problem:
First, even the most complex simulations cannot hope to account for all factors impacting disease progression as it may depend on everything from the weather to shopping habits and from socio-economic status to average household size.
Consequently, there exist a number of hidden variables and effects that are not explicitly modeled.
Second, certain model parameters are expected to be localized with different communities having different starting conditions, i.e. household sizes, economic status, job classifications, etc., as well as different states or even different counties having varying levels of compliance to social distancing or other non-pharmaceutical (NPI) interventions.
Finally, data collection is challenging with often noisy results due to errors, reporting delays, etc., and especially the more sophisticated models are naturally under-constrained as different effects can trade-off against each other, i.e. community transmission may be similarly affected by more mask wearing vs. fewer restaurant visits.

One common approach to compensate for both these challenges and uncertainties is to use a simpler model easier to constrain and fit.
In the case of epidemiological models, this typically means a population based approach like the S(E)IR model~\citep{he2020seir} which expresses the disease progression in terms of {\bf S}usceptible, {\bf E}xposed, {\bf I}infected, and {\bf R}ecovered compartments through a set of coupled ordinary differential equations.
However, these approaches are predominantly phenomological and provide limited insights into the underlying causes.
Furthermore, it is challenging to explore different NPI strategies as none of the important factors, i.e.\ schools, businesses, travel, etc., can be directly modeled. Hence, we focus on more detailed agent-based models~\citep{Germann2006,Halloran2008} which explicitly model individual people, their community characteristics, and interactions.

\noindent In particular, we are using the EpiCast model~\citep{Germann2006} which creates populations of agents based on census track level information and is able to directly model school and workplace closures, household sizes, etc. However, depending on the population size that is modeled and the number of compute cores available, a single simulation may take minutes to hours and thus a brute-force search to match a model to observations is infeasible. Furthermore, some parameters of the model are local, i.e. dependent on the geographical region like the number of currently infected, while others are global in that they describe biological properties of the virus, such as the fraction of asymptomatic infections. Fitting the latter on a per city, county, or even state level may lead to inconsistent results as the fundamental disease parameters are unlikely to differ significantly across regions. Conceptually, the ideal solution to this problem is to construct a nation-wide simulation with consistent global but flexible local parameters. Unfortunately, not only is even a single national simulation computationally expensive but the space of all local parameters could be prohibitively large and this is not yet considering the immense complexity in setting up such a simulation.

In this paper, we consider a more practical formulation of the problem: We develop a single neural network to act as an emulator for \name{} across all geographical regions, using which we jointly estimate local and global parameters simultaneously.  We train this surrogate using simulation runs at the level of Metropolitan Statistical Areas (MSAs), which are densely populated urban centers. By jointly training across MSAs in a population-normalized space (i.e., estimating the fraction of infections with respect to the MSA population), we show that the surrogate generalizes well to previously unseen MSAs. This is particularly useful for modeling MSAs that are infected relatively later in the pandemic and more importantly, it eliminates the need to generate exploratory simulation runs for every MSA. Since inference with a neural network is orders of magnitude cheaper than \name{}, we are able optimize for the optimal set of input parameters required for calibration. In order to make this highly ill-posed inverse problem more tractable, we propose new regularization objectives: (a) explicitly tying the global parameters across geographical regions while optimizing for the optimal set of global and local parameters, and (b) constraining the parameter values to be close to those observed during training. We find that they are effective at constraining the solution space, ultimately yielding accurate calibration of the epidemiological model. We validate our approaches with \name{} and show that the proposed regularizers, when coupled with the pre-trained surrogate, can accurately recover the \name{} parameters required to match the observed data well.

%% file: background.tex
\subsection{\name{}}
\label{sec:epicast}
As discussed above, we are using the \name{} framework as our epidemiological model. Originally, developed to understand influenza outbreaks \citep{Germann2006} it has recently been adapted and modified to model the COVID-19 spread.
\name{} is an agent-based simulation, which explicitly models individual agents in a population, their main occupations and locations, and interactions.
Based on a 12 hour time steps an agent can be at home, at work or school, or in the community representing tasks such as shopping.
Randomized populations are created from census data based on census tract granularity with about 2,000 agents per tract.
\name{} considers a wide range of properties, such as age distributions, house hold sizes, occupation, commuting patterns, etc.
In particular, the system allows epidemiologists to explore detailed scenarios such as different school schedules, the impact of travel restrictions, etc. However, calibrating the model to noisy data especially in the context of shifting public health orders, local variations between populations, and various other confounding factors is difficult.

 \noindent \name takes six parameters to produce the average number of expected infections in the regions of interest. These are: (a) \texttt{INFECTED}: The factor used for the starting number of infected population, (b) \texttt{REMOVED}: percent of the population assumed already removed (quarantined or hospitalized), (c) \texttt{COMPLICANCE} Fraction of people expected to comply with local guidelines, (d) \texttt{TRANSPROP}: transmission probability of the disease, (e) \texttt{PROPASYM}: Fraction of asymptomatic people in the population, and finally (f) \texttt{RELINF}: Relative infectiousness. The first three of these are considered local and can reasonably vary between different regions, while the last three are global as they describe biological properties of the virus.

\noindent We selected a set of self-consistent scenarios to demonstrate how modern surrogate modeling and advanced calibration provide us with highly accurate and globally consistent model parameters across different regions.
In particular, we have created a large ensemble of $15$ metropolitan statistical areas (MSAs), roughly corresponding to major metropolitan areas in the US, aimed to approximate their daily COVID-19 case loads as provided by the Johns Hopkins University (JHU) repository~\citep{dong2020interactive}. Each ensemble explores the six parameters discussed above and is initialized according to the local characteristics of each MSA.
For each MSA, the starting numbers are seeded in an tight interval $\pm 2\%$ of the JHU data assuming a under-reporting of cases by a factor of 3. The latter compensates for the fact that the simulation assumes perfect knowledge of all infections and is inline with best estimates for unreported cases. In separate experiments we have seen limited impact of the exact under-reporting factor and would expect identical results for other choices. All ensembles start at June 22nd and simulate four weeks of disease progression to provide a total of $7,500$ \name{} training curves.

\subsection{Calibration}
Calibration (also called history matching) is the problem of matching the output of a simulation to real observed data. This is often used to better understand the phenomenon of interest. In epidemiology, calibration allows us to determine the state of a pandemic and characterize properties of the infecting agent. Let us define the set of output curves to be $y \in \mathcal{Y} \subset \mathbb{R}^T$, where $T$ is the number of days for which data is available. Next, the inputs are defined as $x \in \mathcal{X}\subset \mathbb{R}^d$, where $d$ is the number of parameters of the epidemiological model; for \name $d=6$. The simulation outputs an expected number of daily infected cases over the period of interest. Now, given an observed curve $y_{obs}\subset \mathbb{R}^t, t\leq T$, the calibration problem is to find the posterior distribution $p(x|y_{obs})$ such that a sample $x$ from this posterior matches the simulation output according to a goodness-of-fit measure (GOF) such as an $\ell_p$ norm. In other words, identifying a set of samples $\mathbf{x} = [x_1,x_2,\dots,x_n], \forall i, x_i \in \mathcal{X}$ such that, $||\mathcal{E}(x_i)- y_{obs}||_p\leq\epsilon$, where $\mathcal{E}$ represents the \name model and $\epsilon$ determines the desired error bound.



%% file: proposed.tex
In this section we outline the proposed surrogate that acts as an emulator for \name in a wide range of MSAs. Next, we describe how this surrogate can be used for calibrating \name in different scenarios.

\subsection{Surrogate training} An accurate neural network based surrogate for the \name simulation can speed up compute by orders of magnitude while producing outputs consistent with \name. As a result, even an approximate surrogate can be very useful for calibration, by enabling several thousand evaluations in order to infer accurate posteriors. In contrast to general surrogate modeling problems, \name has the critical challenge of making consistent predictions across all MSAs -- potentially even ones which may not be accessible during training. This is a practical necessity that can help MSAs that are infected relatively later in the pandemic to leverage data and information from MSAs that are infected earlier.

In order to train such a ``universal'' surrogate to make predictions across MSAs, we need a few simple, reversible data transformations -- (1) first, we consider the number of cumulative cases over time instead of the daily infected cases, since the former is easier to predict (smooth trajectory), (2) we transform the data into a \emph{population normalized space} such that we operate with the number of infections relative to the total population; and (3) finally we reduce the dimensionality of the cumulative case curves using any standard dimensionality reduction method (10-component PCA in our experiments). As a result, surrogate fitting is now reposed as $\hat{\mathcal{E}}:x\mapsto z$, where $z\in \mathbb{R}^{10}$ correspond to the PCA components of the curve; the final predicted curve is obtained by $\hat{y} = \Pi^T z$, where $\Pi$ is the PCA basis estimated using the training data.

\paragraph{Surrogate network architecture} The surrogate is modeled as a fully connected 3-layer neural network that maps a $6$-D input parameter to a $10$-D space using LeakyRelu \citep{xu2015empirical} activations, and batch-normalization \citep{ioffe2015batch}. In general, we find that training in the reduced dimensionality space affords better surrogates with fewer training examples. For more complex curves and larger datasets, $\Pi$ can be replaced with a more sophisticated pre-trained representations, e.g., auto-encoder or a generative model; similar observations have been reported for other surrogate modeling problems in \citep{AnirudhPNAS20}.

\subsection{Surrogate-based calibration of epidemiological models}
Calibration is an inverse problem with a non-linear forward operator, i.e., the epidemiological simulation. Like typical inverse problems, calibration is highly ill-posed, i.e., there are several solutions (input parameters) that may yield the same simulation output. For example, the cumulative number of new infections over the next $T$ days maybe very similar either when the initial number of infections is high and the transmission probability is low or if the number of infections is low but transmission rate is high. As a result, we need to resolve these potential solutions further using prior knowledge of the pandemic, where for instance, an unusually low (or high) transmission rate may not be possible and similarly using information about the current number of infected cases. We formally describe the posterior estimation problem and present the novel regularization strategies next.

\noindent Formally, calibration is the problem of determining $p(x|y_{obs})$, commonly referred to as the posterior distribution, given an observation $y_{obs}$. Here, the observations specify the trajectory of cumulative new infections over the last $T$ days of observation. In addition, $p(y)$ may not be computationally tractable, and as a result determining the posterior in closed form is non-trivial.

\noindent \textbf{Initialization:} The quality of samples obtained is dependent on the initialization for $x$. For example, a relatively safe choice is to draw the initial seed set from an uniform distribution, which is a weak prior since it assumes all values of the parameters are equally likely which matches the desired sampling of the training data. However, given the relatively low number of samples per ensemble (500 in 6-dimensional space) the actual samples produced need not be uniform. Hence, we found it useful to draw $\mathbf{x}_{init}$ from a multivariate normal (MVN) distribution, whose means and covariance are estimated from the input parameter settings in the training set.

\subsubsection{Na\"ive posteriors} The easiest way to estimate the posterior is to choose an appropriate initial set of samples that are expected to cover the range of potential solutions and filter the top-K based on their similarity according to a metric of choice (such as an $\ell_p$-norm). The tightness of the posterior estimate improves with a larger set of initial samples since it increases the odds of finding closer matches. As a result, often requiring tens of thousands of evaluations, which become infeasible with a complex, computationally intensive simulation like \name{}. Instead, here we propose to use the surrogate to evaluate a large number of initial samples ($\sim 200$K) drawn from a multivariate distribution, $\mathbf{x}_{init}$. We then find the top-K matches based only on the data from the observed time-steps, and use that as the approximate posterior. In other words, we find the top-K samples that minimize: $\sum_{i=1}^n ||y_{obs} - \Pi^T(\hat{\mathcal{E}}(x_i))||_p$, where $\|.\|_p$ denotes the $\ell_p$ norm ($p = 2$ in our experiments).

\subsubsection{Optimized posteriors using gradient descent}
The na\"ive approach relies on a large initial set in order to find good matches, which may not always feasible, and requires an exponentially large number of such seed samples to fit a specified error.  Alternatively, we explore a more direct optimization strategy using a much smaller initial seed set (1K samples) and find a better posterior. Here the problem is reformulated as an optimization problem in $x$, and solved using gradient descent. In particular, we can use an optimizer like Adam \citep{kingma2014adam} to search the parameter $\mathcal{X}$ for the optimal $x^*$. This is reminiscent of projected gradient descent (PGD) style optimization strategies that find an appropriate sample in the range space of a pre-trained generator to solve inverse imaging problems~\citep{yeh2017semantic}. For this optimization, we compute the loss directly on the curves, instead of their PCA coefficients, as we find that provides better gradients for our optimization. Mathematically,
\begin{equation}
\label{eq:vanillaGD}
\min_{\{x_i \in \mathcal{X}\}} \left[\sum_{i=1}^n \left\|(y_{obs}- \Pi^T(\hat{\mathcal{E}}(x_i))\right\|_2\right]+\lambda \mathcal{R}(\{x_i\}),
\end{equation}where $\mathcal{R}(.)$ is an appropriate regularization term that reduces the potential solution space. Like in many inverse problems, the choice of the regularizer plays a key role in finding high quality solutions. However, commonly used regularizers such as sparsity or total variation do not apply in the case of epidemiological models since the parameters have a physical meaning (and specific set of ranges) that must be respected.

\paragraph{Regularization objectives for epidemiological model calibration} We consider two different regularization objectives in this work, inspired by the physical quantities represented by the parameters. First, we use a Kullback-Liebler divergence objective between  $\mathbf{x}$ and $\mathbf{x}_{init}$ to ensure that the gradient descent does not ignore the meaningful relationships across input parameters. We approximate $\mathbf{x}$ at each step to be drawn from a multi-variate normal (MVN) distribution, and as a result we can compute the KL divergence in closed form between two MVNs (See \citep{duchi2007derivations} for derivation) as
\begin{equation}
\label{eq:mvn_kld}
\mathrm{KL}(p(x)||p(x_{init})) = \frac{1}{2}\bigg(\log\frac{\det\Sigma_2}{\det\Sigma_1} - d +\mathrm{Tr}(\Sigma_2^{-1}\Sigma_1)+(\mu_2-\mu_1)^T\Sigma_2^{-1}(\mu_2-\mu_1)\bigg)
\end{equation}
where $\mu_1, \mu_2, \Sigma_1, \Sigma_2$ correspond to the mean and covariances for $\mathbf{x}$ and the initialization $\mathbf{x}_{init}$ respectively.
\vspace{5pt}

\noindent\textbf{Consistency conditions for the global parameters}
As discussed above, \name{} uses three local parameters that are expected to vary between regions and three global ones expected to be the same. Therefore, when calibrating the model for a particular region, we can use observations from other regions for which data is available to constrain the unknown global parameters to be consistent. We achieve this by simultaneously optimizing for the parameters across $K$ different MSAs, $\mathbf{x}^{(j)} = \{x_i^{(j)}\} \forall j = 1,\cdots,K$, using the following cost:
\begin{equation}
\mathcal{L}_{multi} = \sum^K_{j=1}\sum_{i=1}^n\|y_{obs}^{(j)}-\Pi^T(\hat{\mathcal{E}}(x_i^{(j)})\|_2,
\end{equation}
We then impose a KL-divergence based cost using only the global parameters, to define this global consistency constraint for each MSA $j$:
\begin{equation}
\label{eq:global}
\mathcal{G}^{(j)} = \sum_{k \neq j}\mathrm{KL}\left(p(x^{(j)})||p(x^{(k)})\right),
\end{equation}where $x^{(j)}$ is the random variable corresponding to the observations in the sample set $\mathbf{x}^{(j)}$. Note, the consistency is imposed across the $K$ different MSAs. The overall optimization objective to calibrate \name{} for all the $K$ different MSAs simultaneously is

\begin{equation}
\label{eq:optcost}
\left\{\mathbf{x}^{(j)*}\right\}_{j=1}^K = \arg \min_{\{\mathbf{x}^{(j)}\}_{j=1}^K} \mathcal{L} =  \mathcal{L}_{multi} +  \lambda_1 \sum_{j=1}^K \mathrm{KL}(p(x^{(j)})||p(x_{init}))  +\lambda_2 \sum_{j=1}^K\mathcal{G}^{(j)},
\end{equation}
where the KL cost is only measured on the $k^{th}$--MSA of interest, while the rest of the terms are used on all the test MSAs available for calibration. $\lambda_1,\lambda_2$ are regularization weights whose values are fixed using cross validation. In all our experiments, we used $\lambda_1 = 1\mathrm{e}-6, \lambda_2=1\mathrm{e}-4$.

\noindent The complete calibration procedure for a single MSA is outlined in algorithm \ref{alg:posterior}, it can be trivially extended to the multi-MSA setting. Following the suggested reporting guideline for calibration methods outlined in \citep{stout2009calibration}, we summarize the key aspects of our calibration strategy: (a) \textit{Target:} Cumulative curve of new infections over a $T$-day period. (b) \textit{GOF metric:} Mean squared error, (c) \textit{Search algorithm:} Gradient descent using a NN-surrogate as outlined in alg. \ref{alg:posterior} with 1000 initial samples. (d) \textit{Acceptance criteria:} Convergence of GOF measure (e) \textit{Stopping rule:} $N_{max} = 25000$ steps.

 \begin{algorithm}[!htb]
	\caption{Proposed Algorithm for Posterior Estimation}
	\label{alg:posterior}
	\begin{algorithmic}[1]
		\Procedure{Calibration}{$y_{obs}^{(j)}, \hat{\mathcal{E}},\Pi, \mathbf{x}_{init}$}
		\State Initialization $\mathbf{x}^{(j)} = \mathbf{x}_{init}, \forall j$
		\For{$n \leftarrow 1~\mathbf{to}~N_{max}$}

			\State $\hat{y}^{(j)} = \Pi^T\hat{\mathcal{E}}(\mathbf{x}^{(j)}), \forall j$ \Comment{$\Pi$: PCA Basis}
			\State
			Compute objection in eqn. \eqref{eq:optcost}
			\State $\mathbf{x}^{(j)}\leftarrow\mathbf{x}^{(j)} - \gamma\nabla\mathcal{L}, \forall j$\Comment{Update estimate }
		\EndFor
			\State \textbf{return} $\{\mathbf{x}^{(j)}\}$ \Comment{samples from the posterior}
		\EndProcedure
	\end{algorithmic}
\end{algorithm}

%% file: expts.tex
\begin{figure}[!htb]
    \centering
    \includegraphics[width=0.7\linewidth]{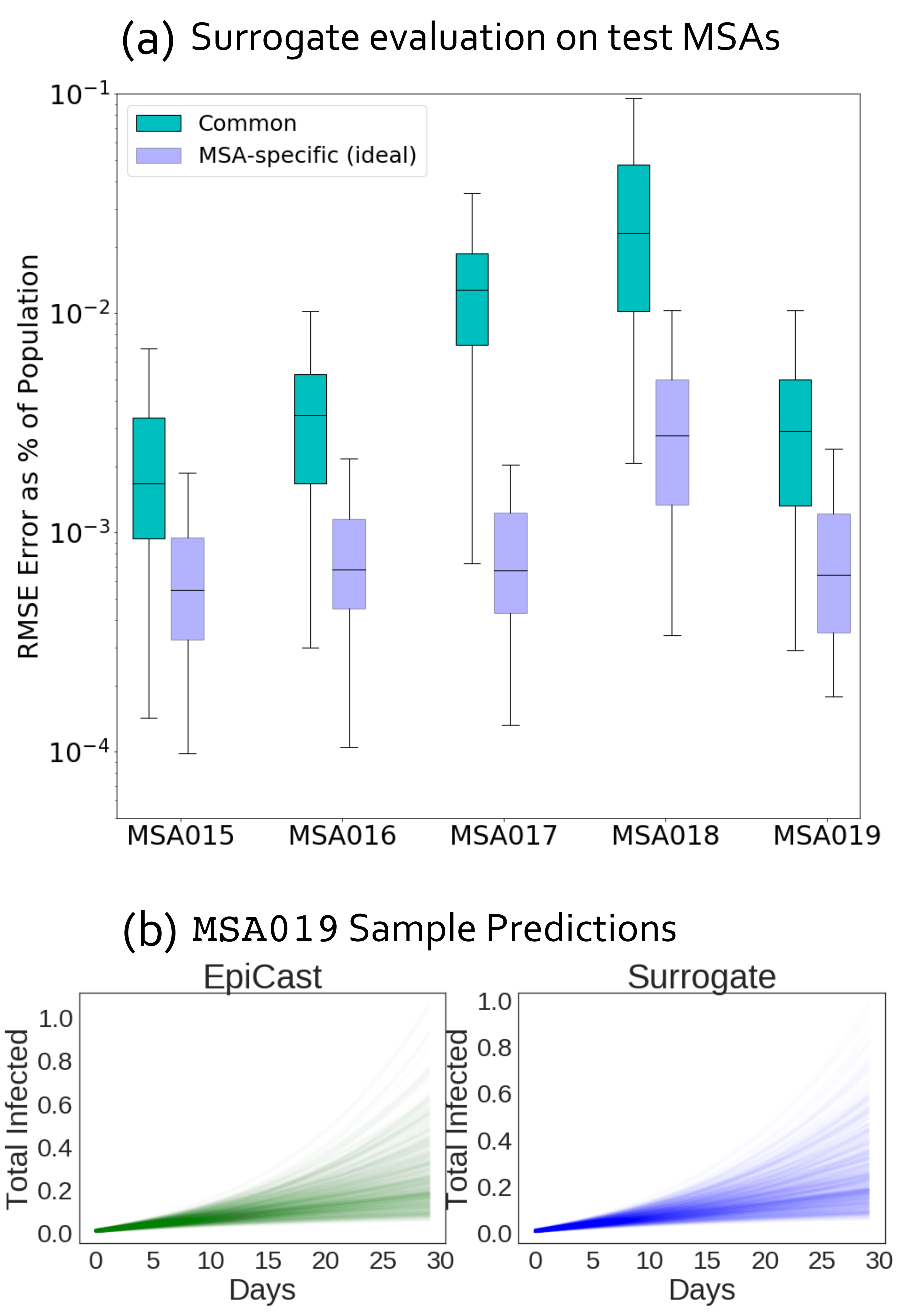}
    \caption{\small{Our \name surrogate can generalize to previously unseen MSAs in population-normalized space, often predicting within $0.1\%$ error of the total population. Here, we show the $25\%$ and $75\%$ confidence intervals estimated using 500 curves across the 5 test-MSAs not accessed during training.}}
    \label{fig:surrogate-eval}
\end{figure}

In this section we outline the experimental details and describe their results. We first describe the training protocol for the surrogate, followed by its use in posterior estimation for calibrating the main epidemiological model.
\subsection{\name{} Simulation Dataset}
As noted in section \ref{sec:epicast}, we create a dataset that is representative of the functionality of \name{} using 7500 runs that consist of 15 MSAs with 500 runs each. The names and descriptions of individual MSAs considered in this work are available in the supplement. Each simulation takes about three minutes on 64 cores of a Knights Landing (KNL) node on one of the top 20 supercomputer. Together with pre- and post-processing simulations, workflow overheads etc., the entire ensemble required $\sim40,000$ core hours. We use 10 MSAs to train our surrogate the remaining for testing. We perform all our calibration experiments using 15 test curves from MSA015.
\begin{figure}[!htb]
    \centering
    \includegraphics[width=0.9\linewidth,clip,trim={25 32 20 20}]{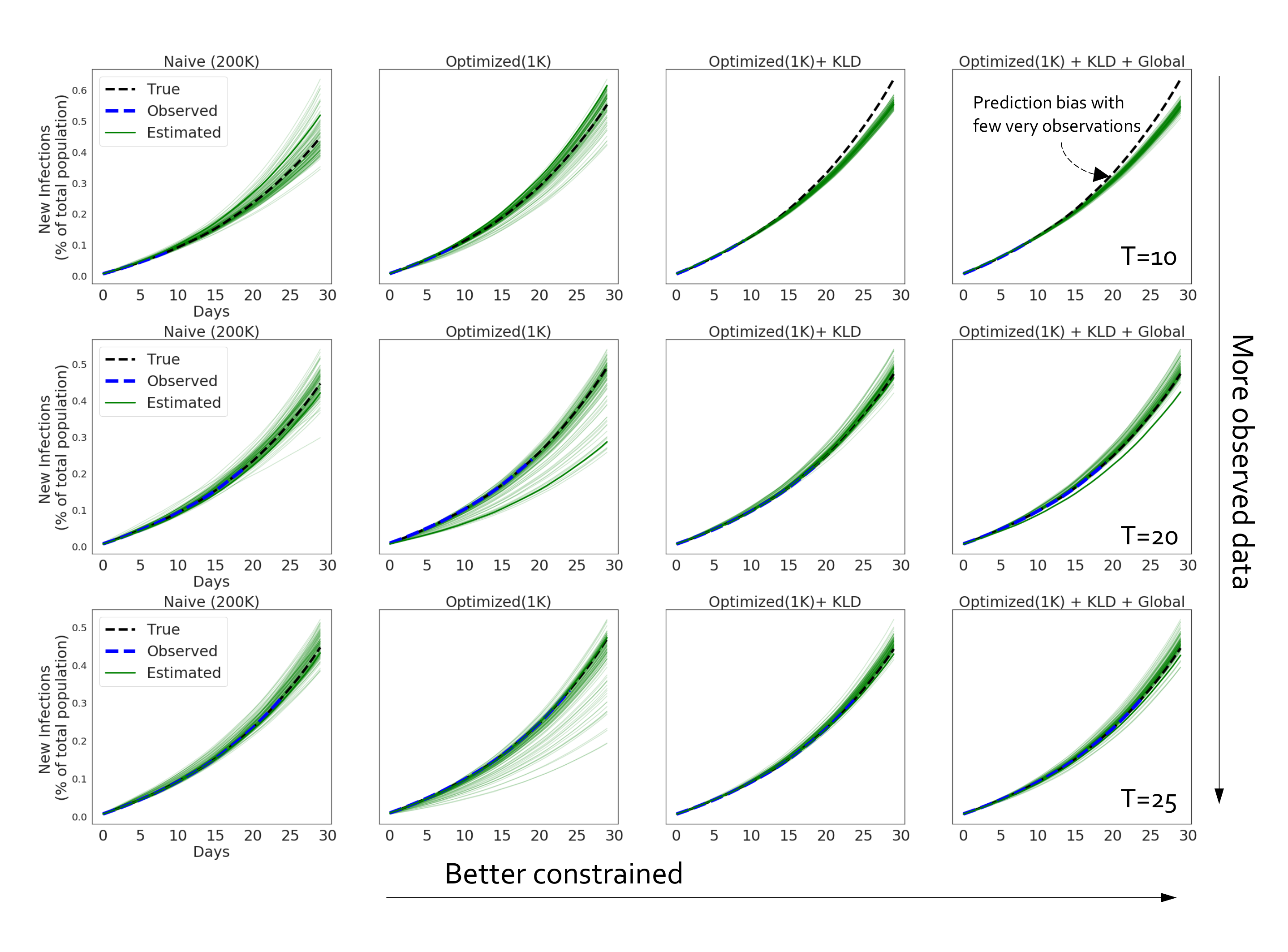}
    \caption{\small{Comparing \name evaluations from the samples obtained using different posterior estimation methods considered here. We show how the estimates change with more data observed.}}
    \label{fig:y_posterior}
\end{figure}

\subsection{Surrogate Training and Evaluation}
The surrogate network consists of a 3 dense layers, along with a LeakyReLu activation and batch normalization after the first and second layers. The surrogate predicts into the PCA latent space, which are decoded using the PCA components to obtain the final curve, as described earlier. We train the model on 500 evaluations of \name{} on 10 MSAs, and test it on 100 curves from 5 test MSAs not accessed during training. Although \name{} produces curves specific to local regions based on demographic factors, we show that a surrogate in population normalized space is able to achieve reasonable performance accuracy. Figure \ref{fig:surrogate-eval}(a) shows the prediction performance using the error as a percentage of the population for each of the 5 test MSAs for both a common surrogate across all MSAs and an MSA-specific surrogate. For the latter, we train on 400 curves from each of the MSAs and evaluate them on the remaining 100 (which are the same test curves as the common model). We observe that in all cases, a surrogate trained on a set of MSAs generalizes well to previously unseen MSAs often with a prediction error of under $0.1\%$ of the total population, indicating similarities in infection spread across dense metropolitan regions. As expected, a ideal surrogate that has seen data from a particular MSA performs better, but not significantly so. In Figure \ref{fig:surrogate-eval}(b) we also show sample predictions from both \name epidemiological model and our NN-surrogate and see that for one of the test MSA019, our predictions match the model accurately, while being more efficient to compute.

\begin{figure}[!htb]
    \centering
    \includegraphics[width=0.98\linewidth]{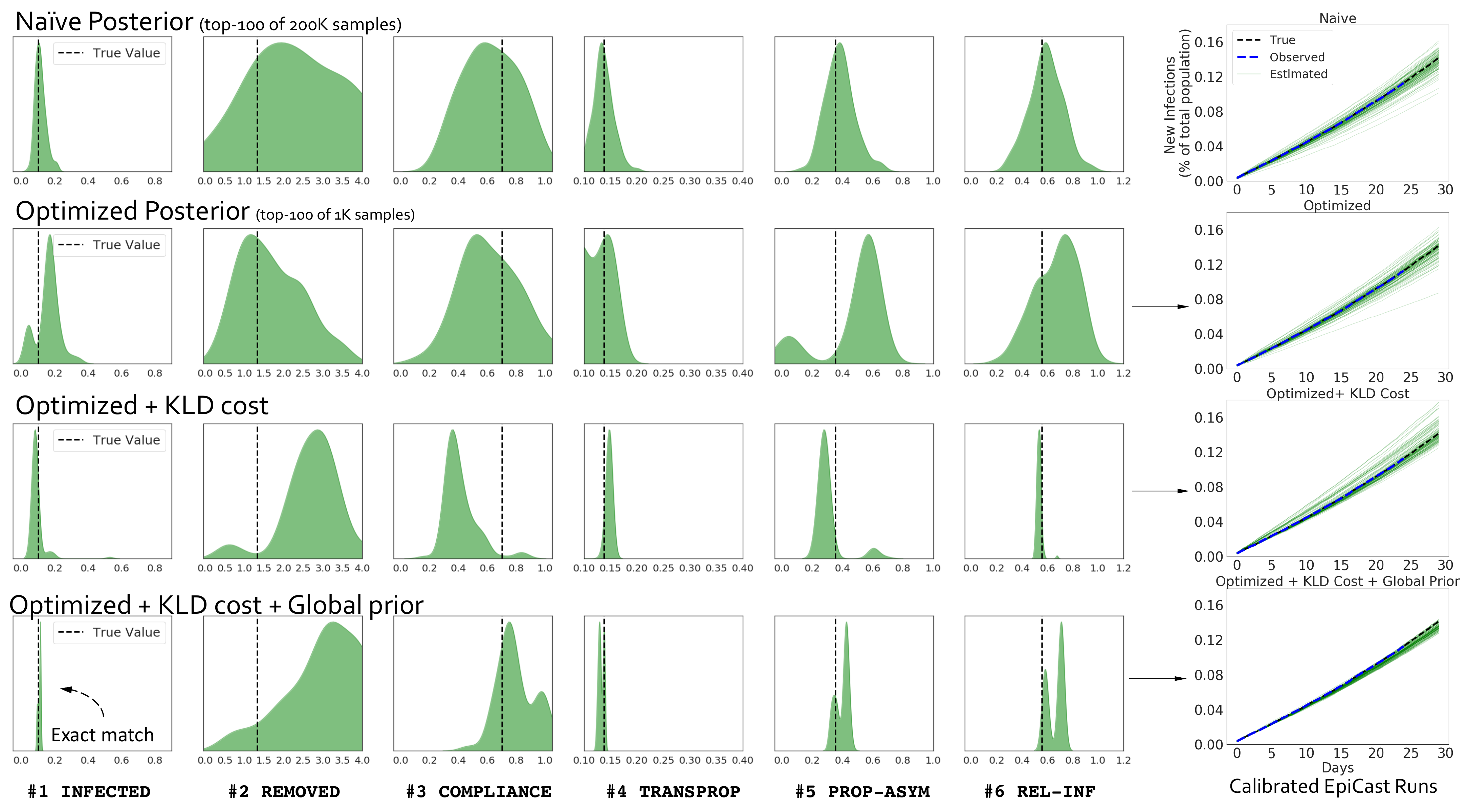}
    \caption{\small{Comparing the marginals of posteriors obtained using different methods outlined in this work. At the bottom, we show the \name evaluations on samples from these posteriors. We observe that the proposed priors yield tighter posteriors, and consequently more accurate matches with the epidemiological model.}}
        \vspace{-10pt}

    \label{fig:x_posterior}
        \vspace{-10pt}

\end{figure}

\begin{table}[!t]
	\centering
	\begin{tabular}{p{0.65in}|p{0.8in}p{0.8in}p{0.8in}|p{0.8in}p{0.8in}p{0.8in}}
		\hline
			\cellcolor{gray!20}\small{\textbf{Method}} & \cellcolor{gray!20} \small{$\mathbf{T_{obs}=10}$} &\cellcolor{gray!20}\small{$\mathbf{T_{obs}=20}$}& \cellcolor{gray!20}\small{$\mathbf{T_{obs}=25}$}&\cellcolor{gray!20}\small{$\mathbf{T_{obs}=10}$}&\cellcolor{gray!20}\small{$\mathbf{T_{obs}=20}$}&\cellcolor{gray!20}\small{$\mathbf{T_{obs}=25}$} \\
			\hdashline
			&  &\textbf{\texttt{INFECTED}} & &&\textbf{\texttt{TRANSPROP}}&\\
		Na\"ive & $0.029\pm 0.023$&$0.029\pm 0.025$& $0.029\pm 0.023$&$0.022\pm 0.015$&$0.021\pm 0.014$ &$0.021\pm 0.014$\\
		Optimized &$0.035\pm 0.032$&$0.063\pm 0.030$&$0.062\pm 0.033$&$0.029\pm 0.020$ &$0.035\pm 0.019$&$0.029\pm 0.018$\\
		+KLD  & $0.015 \pm 0.005$ & $\mathbf{0.025\pm 0.024}$ & $0.024\pm 0.022$&$\mathbf{0.021\pm 0.008}$ &$\mathbf{0.017\pm 0.009}$&$0.017\pm 0.009$\\
		+Global &$\mathbf{0.014\pm 0.006}$ & $0.028\pm0.022$ & $\mathbf{0.023\pm0.014}$&$\mathbf{0.021\pm 0.009}$&$\mathbf{0.017\pm 0.008}$&$\mathbf{0.014\pm 0.006}$\\
			\hline
	\end{tabular}
		\caption{Evaluating RMSE on the two most sensitive parameters \texttt{INFECTED, TRANSPROP}. We observe that the additional constraints studied in this paper improve the quality of the estimate. The mean prediction and standard deviation are averaged across 20 different observations from a test set corresponding to MSA015.}
	\label{tab:x_rmse}
\end{table}

\subsection{Calibration and posterior estimation}
To evaluate calibration performance we use 15 curves from MSA015, which is one of the given test MSAs. We use the inputs that were originally used to generate these curves using \name{} as the ``true values'' in our evaluations. Since the problem is highly ill-posed we expect several combinations of the same parameters to produce the same curve, as a result, we evaluate the performance in two quantitative ways (a) a root-mean squared error (RMSE) goodness of fit  compared to the observed curve and those produced by evaluating \name on samples from the posterior for each method in consideration. (b) We also measure the RMSE of the estimated parameters on the two most sensitive parameters directly -- \texttt{INFECTED} and \texttt{TRANSPROP} which correspond to the number of infected at the start of measurements, and the transmission probability of the infections respectively. In reality, the global consistency cost can be directly applied across MSAs, but since we are working with simulated data, we find samples across the test MSAs with similar global parameters and use the corresponding curves across MSA 015, 016, 017 in computing the cost according to \eqref{eq:global}.

\begin{table}[!htb]
	\centering
	\begin{tabular}{p{0.5in}ccc}
		\hline
		\small
			\cellcolor{gray!20}\textbf{Method} & \cellcolor{gray!20}$\mathbf{T_{obs}=10}$& \cellcolor{gray!20}$\mathbf{T_{obs}=20}$ & \cellcolor{gray!20}$\mathbf{T_{obs}=25}$ \\
			\hdashline
			Na\"ive & $9.93 \pm 6.88$& $7.07\pm 5.61$& $6.70 \pm 5.57$ \\
			Optimized &$\mathbf{8.93\pm 9.11}$ &$16.64\pm 17.17$ & $13.06\pm 12.17$\\
			+ KLD  & $11.42 \pm 2.38$ & $\mathbf{5.20\pm 4.59}$ & $5.47\pm4.81$ \\
			+ Global &$11.63 \pm 2.55$ & $6.23\pm5.26$ & $\mathbf{5.42\pm4.37}$\\
			\hline
	\end{tabular}
		\caption{Evaluating RMSE quality (as percentage of population $\times 1\mathrm{e}-3)$ for curve fits by running samples by evaluating samples from the posterior with the \name simulation.}
\label{table:main_Expt}
\end{table}
\subsubsection{Qualitative results} In Figure \ref{fig:y_posterior} we show how the quality of calibration changes with more available data, and better constraints. In each case we run \name on the samples generated and show the observed curve, the true curve, and the curves produced by the simulation. Ideally, we want the curves from the simulation to be as close as possible to the true curve, given only the observed curve. As we observe in Fig \ref{fig:y_posterior}, when very few observations are available the na\"ive model and the least constrained optimized posterior are most likely to contain the true solution, mainly because these methods have a very large variance and their posteriors are weak. Next, as we observe more data, we observe that including the proposed regularization objectives significantly improves the quality of the fits for the same number of \name evaluations.

\noindent In Figure \ref{fig:x_posterior}, we show the marginal distribution of samples from the posterior distribution on all six parameters -- the first three are ``local'' parameters, i.e., specific to the geographical region and the last three are ``global''. We compare all four methods on a test curve from RMSA015, with respect to the true value which is shown as a black dashed line. As expected, we observe that the marginals for samples obtained using the naive and the unconstrained optimization methods yield very wide posteriors, even on highly sensitive parameters like \texttt{INFECTED} and \texttt{TRANSPROP}, that are expected to be tight. On the other hand, the samples obtained using the regularized optimization yield very confident estimates for all the parameters. We also show the corresponding \name evaluations of these samples and observe that the proposed constraints significantly help improve the posterior fits, yielding accurate curves compared to the ground truth.

\subsubsection{Quantitative evaluation}
\noindent In table \ref{table:main_Expt}, we show the RMSE error for the top-100 curves (based on GOF), across 15 test curves from RMSA015. For each test curve, we estimate the samples from the posterior followed by evaluating the top 100 sample candidates  using \name. In all cases, the cost for optimization or sample selection/filtering is done based on curves for the number of days they can be observed. Next, in table \ref{tab:x_rmse} we evaluate the quality of estimating the two mose sensitive parameters to \name. Since the problem of calibration is highly ill-posed, average metrics comparing samples from the posterior distribution to the ground truth may not be directly meaningful, however the sensitive parameters are expected to be easier to estimate (as seen in Fig \ref{fig:y_posterior}). We observe that in all scenarios considered across both the parameters, the regularization objectives help in optimization and in the accurate recovery of sensitive parameters.

%% file: related.tex
Agent-based models (ABMs)~\citep{perez2009agent,Germann2006} are computational models that are commonly used to inform policy-making regarding public health. Such models must be `calibrated` in order to explain observed data, which involves identifying the input parameters to the computational model that can exactly match the observed data according to a goodness of fit (GOF) metric. Since this process is under specified, there is often a large set of solutions that yield the same output to match the observation. Within epidemiology there are several studies that address the calibration problem for both agent-based and SEIR models \citep{dantas2018calibration,ward2016dynamic,farah2014bayesian}, including numerous recent works related to the COVID-19 pandemic\citep{bertozzi2020challenges,hazelbag2020calibration}. The focus of our work is specifically on ABMs, where previous works have mostly focused on calibration achieved via primarily parameter searches or optimization strategies based on GOF metrics such as R-squared, RMSE, absolute error etc. We refer the interested reader to Tables 1\&2 of \citep{hazelbag2020calibration} for a comprehensive list of these works. 

\noindent Most existing works have either focused on optimization strategies to obtain point estimates, or sampling algorithms to find a distribution of parameter values to approximate the posterior distribution. In contrast, our approach relies on a non-linear neural network-based surrogate that acts as a proxy for the computationally intensive epidemiological model. Since inference with a neural network is orders of magnitude cheaper than the original simulation, we are able to employ complex optimization strategies for calibration. To the best of our knowledge we are the first to propose calibration using a neural network based surrogate for ABMs. Our work is also related to an increasing number of covid modeling efforts using machine learning \citep{davis2020use,soures2020sirnet,libin2020deep,Covid19ML} that have been used to model and design mitigations with simpler epidemiological models like the SEIR models. In contrast, we are interested in the more complex, computationally intensive ABMs. Similar to our work,~\citep{davis2020use} focus on the specific problem of surrogate modeling for SIR models and other individual-based models. Our focus is on building an effective surrogate that can also be used in calibration across geographical regions not accessed during training.

%% file: discussion.tex
Epidemiological models play a crucial role in assessing the state of a public health crisis like the ongoing COVID-19 pandemic. These computational models enable decision makers to explore what-if scenarios and forecasts acting as an important planning tool for implementing health policies nationwide. Like all computational models in order for them to be effective, they must be calibrated to fit observed data, which is the process of finding the right combination of input parameters that can match the observed data to some desired degree of fidelity. We are specifically interested in calibration of complex agent-based models (ABMs), that are computationally intensive. We propose an alternative strategy for using an ABM called \name{} \citep{Germann2006} with the help of a neural network-based surrogate. We demonstrate that our surrogate can effecitvely emulate the ABM's behaviour -- including across geographical regions that were not included in the training dataset. We additionally propose novel regularization objectivesto make the highly ill-posed inverse problem more tractable, and can successfully matches observed curves when samples from our estimated posteriors are evaluated using\name{}. We expect calibration strategies like those proposed in this work to help in more effective usage of ABMs in characterizing and fighting the spread of COVID-19.

%% file: paper.bbl
\begin{thebibliography}{21}
\providecommand{\natexlab}[1]{#1}
\providecommand{\url}[1]{\texttt{#1}}
\expandafter\ifx\csname urlstyle\endcsname\relax
  \providecommand{\doi}[1]{doi: #1}\else
  \providecommand{\doi}{doi: \begingroup \urlstyle{rm}\Url}\fi

\bibitem[Anirudh et~al.(2020)Anirudh, Thiagarajan, Bremer, and
  Spears]{AnirudhPNAS20}
Rushil Anirudh, Jayaraman~J. Thiagarajan, Peer-Timo Bremer, and Brian~K.
  Spears.
\newblock Improved surrogates in inertial confinement fusion with manifold and
  cycle consistencies.
\newblock \emph{Proceedings of the National Academy of Sciences}, 117\penalty0
  (18):\penalty0 9741--9746, 2020.
\newblock ISSN 0027-8424.
\newblock \doi{10.1073/pnas.1916634117}.
\newblock URL \url{https://www.pnas.org/content/117/18/9741}.

\bibitem[Bertozzi et~al.(2020)Bertozzi, Franco, Mohler, Short, and
  Sledge]{bertozzi2020challenges}
Andrea~L Bertozzi, Elisa Franco, George Mohler, Martin~B Short, and Daniel
  Sledge.
\newblock The challenges of modeling and forecasting the spread of covid-19.
\newblock \emph{arXiv preprint arXiv:2004.04741}, 2020.

\bibitem[Dantas et~al.(2018)Dantas, Tosin, and Cunha~Jr]{dantas2018calibration}
Eber Dantas, Michel Tosin, and Americo Cunha~Jr.
\newblock Calibration of a seir--sei epidemic model to describe the zika virus
  outbreak in brazil.
\newblock \emph{Applied Mathematics and Computation}, 338:\penalty0 249--259,
  2018.

\bibitem[Davis et~al.(2020)Davis, Hollingsworth, Caudron, and
  Irvine]{davis2020use}
Christopher~N Davis, T~Deirdre Hollingsworth, Quentin Caudron, and Michael~A
  Irvine.
\newblock The use of mixture density networks in the emulation of complex
  epidemiological individual-based models.
\newblock \emph{PLoS computational biology}, 16\penalty0 (3):\penalty0
  e1006869, 2020.

\bibitem[Dong et~al.(2020)Dong, Du, and Gardner]{dong2020interactive}
Ensheng Dong, Hongru Du, and Lauren Gardner.
\newblock An interactive web-based dashboard to track covid-19 in real time.
\newblock \emph{The Lancet infectious diseases}, 20\penalty0 (5):\penalty0
  533--534, 2020.

\bibitem[Duchi(2007)]{duchi2007derivations}
John Duchi.
\newblock Derivations for linear algebra and optimization.
\newblock \emph{Berkeley, California}, 3:\penalty0 2325--5870, 2007.

\bibitem[Farah et~al.(2014)Farah, Birrell, Conti, and
  Angelis]{farah2014bayesian}
Marian Farah, Paul Birrell, Stefano Conti, and Daniela~De Angelis.
\newblock Bayesian emulation and calibration of a dynamic epidemic model for
  a/h1n1 influenza.
\newblock \emph{Journal of the American Statistical Association}, 109\penalty0
  (508):\penalty0 1398--1411, 2014.

\bibitem[Germann et~al.(2006)Germann, Kadau, Longini, and Macken]{Germann2006}
Timothy~C. Germann, Kai Kadau, Ira~M. Longini, and Catherine~A. Macken.
\newblock Mitigation strategies for pandemic influenza in the united states.
\newblock \emph{Proceedings of the National Academy of Sciences}, 103\penalty0
  (15):\penalty0 5935--5940, 2006.
\newblock \doi{10.1073/pnas.0601266103}.
\newblock URL \url{https://www.pnas.org/content/103/15/5935}.

\bibitem[Gu(2020)]{Covid19ML}
Youyang Gu.
\newblock Covid-19 projections using machine learning.
\newblock \url{https://covid19-projections.com/}, 2020.
\newblock Accessed: 2020-Sept-09.

\bibitem[Halloran et~al.(2008)Halloran, Ferguson, Eubank, Longini, Cummings,
  Lewis, Xu, Fraser, Vullikanti, Germann, Wagener, Beckman, Kadau, Barrett,
  Macken, Burke, and Cooley]{Halloran2008}
M.~Elizabeth Halloran, Neil~M. Ferguson, Stephen Eubank, Ira~M. Longini, Derek
  A.~T. Cummings, Bryan Lewis, Shufu Xu, Christophe Fraser, Anil Vullikanti,
  Timothy~C. Germann, Diane Wagener, Richard Beckman, Kai Kadau, Chris Barrett,
  Catherine~A. Macken, Donald~S. Burke, and Philip Cooley.
\newblock Modeling targeted layered containment of an influenza pandemic in the
  united states.
\newblock \emph{Proceedings of the National Academy of Sciences}, 105\penalty0
  (12):\penalty0 4639--4644, 2008.
\newblock ISSN 0027-8424.
\newblock \doi{10.1073/pnas.0706849105}.
\newblock URL \url{https://www.pnas.org/content/105/12/4639}.

\bibitem[Hazelbag et~al.(2020)Hazelbag, Dushoff, Dominic, Mthombothi, and
  Delva]{hazelbag2020calibration}
C~Marijn Hazelbag, Jonathan Dushoff, Emanuel~M Dominic, Zinhle~E Mthombothi,
  and Wim Delva.
\newblock Calibration of individual-based models to epidemiological data: A
  systematic review.
\newblock \emph{PLoS computational biology}, 16\penalty0 (5):\penalty0
  e1007893, 2020.

\bibitem[He et~al.(2020)He, Peng, and Sun]{he2020seir}
Shaobo He, Yuexi Peng, and Kehui Sun.
\newblock Seir modeling of the covid-19 and its dynamics.
\newblock \emph{Nonlinear Dynamics}, pages 1--14, 2020.

\bibitem[Ioffe and Szegedy(2015)]{ioffe2015batch}
Sergey Ioffe and Christian Szegedy.
\newblock Batch normalization: Accelerating deep network training by reducing
  internal covariate shift.
\newblock \emph{arXiv preprint arXiv:1502.03167}, 2015.

\bibitem[Kingma and Ba(2014)]{kingma2014adam}
Diederik~P Kingma and Jimmy Ba.
\newblock Adam: A method for stochastic optimization.
\newblock \emph{arXiv preprint arXiv:1412.6980}, 2014.

\bibitem[Libin et~al.(2020)Libin, Moonens, Verstraeten, Perez-Sanjines, Hens,
  Lemey, and Now{\'e}]{libin2020deep}
Pieter Libin, Arno Moonens, Timothy Verstraeten, Fabian Perez-Sanjines, Niel
  Hens, Philippe Lemey, and Ann Now{\'e}.
\newblock Deep reinforcement learning for large-scale epidemic control.
\newblock \emph{arXiv preprint arXiv:2003.13676}, 2020.

\bibitem[Perez and Dragicevic(2009)]{perez2009agent}
Liliana Perez and Suzana Dragicevic.
\newblock An agent-based approach for modeling dynamics of contagious disease
  spread.
\newblock \emph{International journal of health geographics}, 8\penalty0
  (1):\penalty0 50, 2009.

\bibitem[Soures et~al.(2020)Soures, Chambers, Carmichael, Daram, Shah, Clark,
  Potter, and Kudithipudi]{soures2020sirnet}
Nicholas Soures, David Chambers, Zachariah Carmichael, Anurag Daram, Dimpy~P
  Shah, Kal Clark, Lloyd Potter, and Dhireesha Kudithipudi.
\newblock Sirnet: Understanding social distancing measures with hybrid neural
  network model for covid-19 infectious spread.
\newblock \emph{arXiv preprint arXiv:2004.10376}, 2020.

\bibitem[Stout et~al.(2009)Stout, Knudsen, Kong, McMahon, and
  Gazelle]{stout2009calibration}
Natasha~K Stout, Amy~B Knudsen, Chung~Yin Kong, Pamela~M McMahon, and G~Scott
  Gazelle.
\newblock Calibration methods used in cancer simulation models and suggested
  reporting guidelines.
\newblock \emph{Pharmacoeconomics}, 27\penalty0 (7):\penalty0 533--545, 2009.

\bibitem[Ward et~al.(2016)Ward, Evans, and Malleson]{ward2016dynamic}
Jonathan~A Ward, Andrew~J Evans, and Nicolas~S Malleson.
\newblock Dynamic calibration of agent-based models using data assimilation.
\newblock \emph{Royal Society open science}, 3\penalty0 (4):\penalty0 150703,
  2016.

\bibitem[Xu et~al.(2015)Xu, Wang, Chen, and Li]{xu2015empirical}
Bing Xu, Naiyan Wang, Tianqi Chen, and Mu~Li.
\newblock Empirical evaluation of rectified activations in convolutional
  network.
\newblock \emph{arXiv preprint arXiv:1505.00853}, 2015.

\bibitem[Yeh et~al.(2017)Yeh, Chen, Yian~Lim, Schwing, Hasegawa-Johnson, and
  Do]{yeh2017semantic}
Raymond~A Yeh, Chen Chen, Teck Yian~Lim, Alexander~G Schwing, Mark
  Hasegawa-Johnson, and Minh~N Do.
\newblock Semantic image inpainting with deep generative models.
\newblock In \emph{Proceedings of the IEEE conference on computer vision and
  pattern recognition}, pages 5485--5493, 2017.

\end{thebibliography}
